# Robot-Assisted Surgical Training Over Several Days in a Virtual Surgical Environment with Divergent and Convergent Force Fields

Y. A. Oquendo[1], Z. Chua[1], M. M. Coad[1], I. Nisky[2], A. Jarc[3],
S. Wren[1], T.S. Lendvay[4], A. M. Okamura[1]

[1] Stanford University, [2] Ben-Gurion University of the Negev,
[3] Intuitive Surgical, [4] University of Washington (oquendoy@stanford.edu)

## INTRODUCTION

Surgical procedures require a high level of technical skill to ensure efficiency and patient safety. Due to the direct effect of surgeon skill on patient outcomes, the development of cost-effective and realistic training methods is imperative to accelerate skill acquisition [1]. Teleoperated robotic devices allow for intuitive ergonomic control, but the learning curve for these systems remains steep. Recent studies in motor learning have shown that visual or physical exaggeration of errors helps trainees to learn to perform tasks faster and more accurately [2]. In this study, we extended the work from two previous studies [3,4] to investigate the performance of subjects in different force field training conditions, including convergent (assistive), divergent (resistive), and no force field (null). We hypothesized that the group who trained in the divergent force field would show superior performance to those in the null and convergent fields.

## MATERIALS AND METHODS

*Experimental setup and task*: We used the da Vinci Research Kit (dVRK) [5] with the Assisted Teleoperation with Augmented Reality (ATAR) framework [4] to create a virtual reality environment for a ring-on-wire manipulation task. The robotic tools were simulated as kinematic objects resembling da Vinci endoscopic mega needle drivers. Subjects in the convergent (C) field received an assistive force that pushed the participant's hand towards an ideal path, while those in the divergent (D) field received a resistive force that pushed the participant's hand away from the ideal path. In both groups, the force applied had a magnitude proportional to their distance away from the path. Force feedback was only applied when the gripper was closed on the ring that was being manipulated. Given the current pose ($P_C = [T_C, Q_C]$) and the desired pose ($P_D = [T_D, Q_D]$), where $T_X$ is a 3-DOF position vector and $Q_X$ is a 4-DOF unit quaternion, the force ($\mathbf{F}$) and torque ($\boldsymbol{\tau}$) produced on the tip of each instrument were:

$$\mathbf{F} = -k_T*(\mathbf{T}_C - \mathbf{T}_D) - d_T*\dot{\mathbf{T}}_C$$
$$\boldsymbol{\tau} = -k_R*rpy(\mathbf{Q}_D*\mathbf{Q}_C^{-1}) - d_R*\boldsymbol{\omega}_C,$$

where $\dot{\mathbf{T}}_C$ and $\boldsymbol{\omega}_C$ are the current translational and angular rate, respectively, $k_T$ and $k_R$ are the translational and rotational proportional coefficients, $d_T$ and $d_R$ are the translational and rotational damping coefficients, and rpy() represents a transformation from quaternion to roll-pitch-yaw. $k_T$, $k_R$ were positive in the convergent field and negative in the divergent field. $d_T$ and $d_R$ were positive in both force fields.

The visuomotor task adapted from [4] was an adaptation of the '*steady-hand*' game often included in surgical training curricula [6]. The user moves a ring along a curved wire which spans along all three Cartesian dimensions and requires up to 90° wrist rotations from each hand for completion. Users received visual feedback on their position such that the ring's color shifted gradually from red (less accurate) to yellow (more accurate) as the ring approached the wire.

*Procedure*: 40 surgical novices gave informed consent for the IRB-approved study, and were pseudo-randomized to one of three groups: convergent force field (n=13), divergent force field (n=15), or null field (n=12). Two divergent field subjects dropped out of the study due to physical limitations preventing task completion, and their data was not included in the analysis. Subjects were required to participate over five consecutive days, and were given a $75 Amazon gift card as compensation. On the day 1, subjects were consented, shown a video demo of the steady-hand task, and directed on how to use the dVRK system. They were instructed to perform the task as accurately and quickly as possible, with more emphasis being placed on accuracy. Subjects then completed 5 baseline trials in the null field followed by 15 trials in their assigned force field. On days 2, 3, and 4, subjects completed 20 trials per day in their assigned force field. On day 5, subjects completed 20 trials in the null field to evaluate changes in performance from baseline. A total of 20 trials per day was chosen to minimize the effect of fatigue from confounding results, and a total of five days was chosen to correspond with a work week for ease of scheduling.

*Data Analysis*: Data was recorded at 30 Hz. Four performance metrics were calculated for each trial: trial time (Time [s]), translational path error (TPE [mm$^2$]) and rotational path error (RPE [Rad·mm]) were calculated as in [3], and combined error-time (CET [Rad·mm·s]) was calculated as follows:

$$CET = Time \cdot (RPE + cf \cdot TPE)$$

Where cf is a constant factor of 17 rad/mm derived from the ratio of the average rotational path error and the average translational path error across all subjects and all trials. Improvement was calculated for each metric as the difference between the mean on the final day and the mean at baseline. Metrics were non-normally distributed between groups, thus we used non-parametric statistical tests to evaluate differences in performance. Kruskal-Wallis (KW) tests were performed with the different metrics as dependent variables and the training group as independent factor. Post-hoc testing was done using the Dunn Test (DT) to determine significant differences between pairs of groups. Pairwise Wilcoxon signed-rank tests (WSR) were used to determine significance of within-subjects improvement.

**RESULTS**

Figure 1 presents the 25$^{th}$, 50$^{th}$, and 75$^{th}$ percentiles of performance of each group at baseline and on the final day of evaluation. At baseline, the groups significantly differed in time to completion (KW; $\chi^2$=7.038, df=2, p=0.030). The null group had significantly lower time to completion (DT; Z=2.51, p=0.036). There was no significant difference in TPE, RPE, or CET at baseline.

The groups significantly differed in CET on the final day (KW; $\chi^2$=6.8501, df=2, p=0.032). Further analysis revealed a significant difference between convergent and null groups (DT; Z=2.57, p=0.030). Divergent field subjects had the lowest final day median combined error-time, though this was not statistically significant. There were no significant differences between groups in any other metric.

Combined performance variability (CPV), measured as the standard deviation of CET across all trials, significantly decreased from baseline in all groups (WSR; C: p=0.017, D: p<0.01, N: p<0.01). The groups differed significantly in their CPV on the final day (KW; $\chi^2$=7.8415, df=2, p-value=0.020), and a Dunn test for that day showed that convergent subjects had significantly higher variability than null subjects (DT; Z=2.77, p=0.016).

Figure 2 shows that all groups significantly improved their performance from baseline. Compared to the null and convergent groups, divergent subjects experienced the most improvement in time (KW; $\chi^2$=3.18, df=2, p=0.20), TPE (KW; $\chi^2$=1.80, df=2, p=0.41), RPE (KW; $\chi^2$=2.37, df=2, p=0.30), and CET (KW; $\chi^2$=3.37, df=2, p=0.18).

**DISCUSSION**

Divergent field subjects improved the most from baseline in all metrics. The difference between groups was not statistically significant, possibly due to outliers in the convergent field and null field groups. Even though the divergent group was significantly slower to complete the task than the null group at baseline, there was no significant difference on final evaluation.

The most significant difference between groups on the final day was in CET, which takes into account the speed-accuracy tradeoff. In surgery, the balance between speed and accuracy is especially important; it is appropriate to move quickly when the stakes of a surgical step are low, and slowly and accurately during the steps where an error could be catastrophic.

Surprisingly, even after four days of training, participants did not reach a plateau in their performance (learning curves not shown). We hypothesize that with more days to train or more trials per day, these differences may become more apparent.

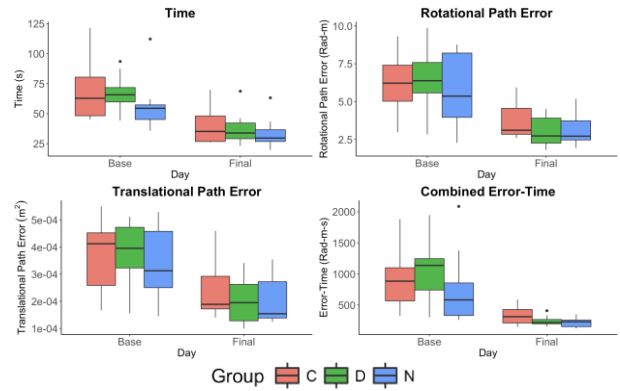

**Figure. 1** The 25th, 50th (black line), and 75th percentile of each metric for each group at baseline and final evaluation. Black dots represent outliers.

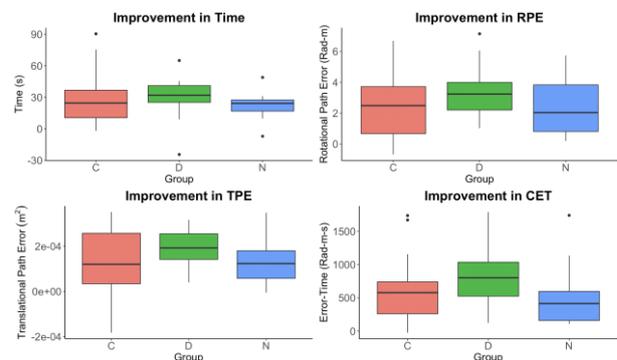

**Figure. 2** Improvements from Baseline (25th, 50th, and 75th percentiles). Black dots represent outliers. All values had p < 0.01 for intragroup difference using a two-sided, paired analysis with Wilcoxon signed-rank test.